\documentclass[letterpaper]{article} 
\usepackage{aaai24}  
\usepackage{times}  
\usepackage{helvet}  
\usepackage{courier}  
\usepackage[hyphens]{url}  
\usepackage{graphicx} 
\usepackage{natbib}  
\usepackage{caption} 
\usepackage{algorithm}
\usepackage{algorithmic}

\usepackage{xcolor}
\definecolor{darkgreen}{rgb}{0,0.5,0}

\usepackage{amsmath} 
\usepackage{amsfonts} 
\usepackage{amssymb} 
\usepackage{bm}
\usepackage{multirow}
\usepackage{makecell}
\usepackage{graphicx}
\usepackage{booktabs}

\newcommand{\x}{{\bf x}}

\usepackage{newfloat}
\usepackage{listings}
\DeclareCaptionStyle{ruled}{labelfont=normalfont,labelsep=colon,strut=off} 
\floatstyle{ruled}
\newfloat{listing}{tb}{lst}{}
\floatname{listing}{Listing}
\title{Open-Set Facial Expression Recognition}

\author{Yuhang Zhang\textsuperscript{\rm 1}, Yue Yao\textsuperscript{\rm 2}, Xuannan Liu\textsuperscript{\rm 1}, Lixiong Qin\textsuperscript{\rm 1}, Wenjing Wang\textsuperscript{\rm 1}, Weihong Deng\textsuperscript{\rm 1}}

\affiliations{
    \textsuperscript{\rm 1}Beijing University of Posts and Telecommunications\\
    \textsuperscript{\rm 2}Australian National University\\

    \{zyhzyh, liuxuannan, lxqin, wwj311, whdeng\}@bupt.edu.cn, yue.yao@anu.edu.au

}

\usepackage{bibentry}

\begin{document}

\maketitle

\begin{abstract}

Facial expression recognition (FER) models are typically trained on datasets with a fixed number of seven basic classes. However, recent research works~\cite{cowen2021sixteen, bryant2022multi, kollias2023multi} point out that there are far more expressions than the basic ones. Thus, when these models are deployed in the real world, they may encounter unknown classes, such as compound expressions that cannot be classified into existing basic classes. To address this issue, we propose the open-set FER task for the first time. Though there are many existing open-set recognition methods, we argue that they do not work well for open-set FER because FER data are all human faces with very small inter-class distances, which makes the open-set samples very similar to close-set samples. In this paper, we are the first to transform the disadvantage of small inter-class distance into an advantage by proposing a new way for open-set FER. Specifically, we find that small inter-class distance allows for sparsely distributed pseudo labels of open-set samples, which can be viewed as symmetric noisy labels. Based on this novel observation, we convert the open-set FER to a noisy label detection problem. We further propose a novel method that incorporates attention map consistency and cycle training to detect the open-set samples. Extensive experiments on various FER datasets demonstrate that our method clearly outperforms state-of-the-art open-set recognition methods by large margins. Code is available at https://github.com/zyh-uaiaaaa.
\end{abstract}

\section{Introduction}
\label{sec:intro}

Facial expression recognition (FER) is vital in human-centered computing as it helps machines understand human feelings~\cite{li2020deep}. Existing FER models are trained with the fixed seven basic classes. However, as pointed out by recent research works published in Nature and top computer vision conferences~\cite{cowen2021sixteen, bryant2022multi, kollias2023multi}, humans can display various expressions that go beyond the basic classes in real-world deployments, such as other different expressions and compound expressions. Close-set FER models trained on the basic classes are unreliable when encountering new unknown expressions as these samples are always misclassified as one of the given classes with high confidence. This limitation hinders the real-world deployment of FER models.

\begin{figure}[t]
  \centering
  \includegraphics[width=1.\linewidth]{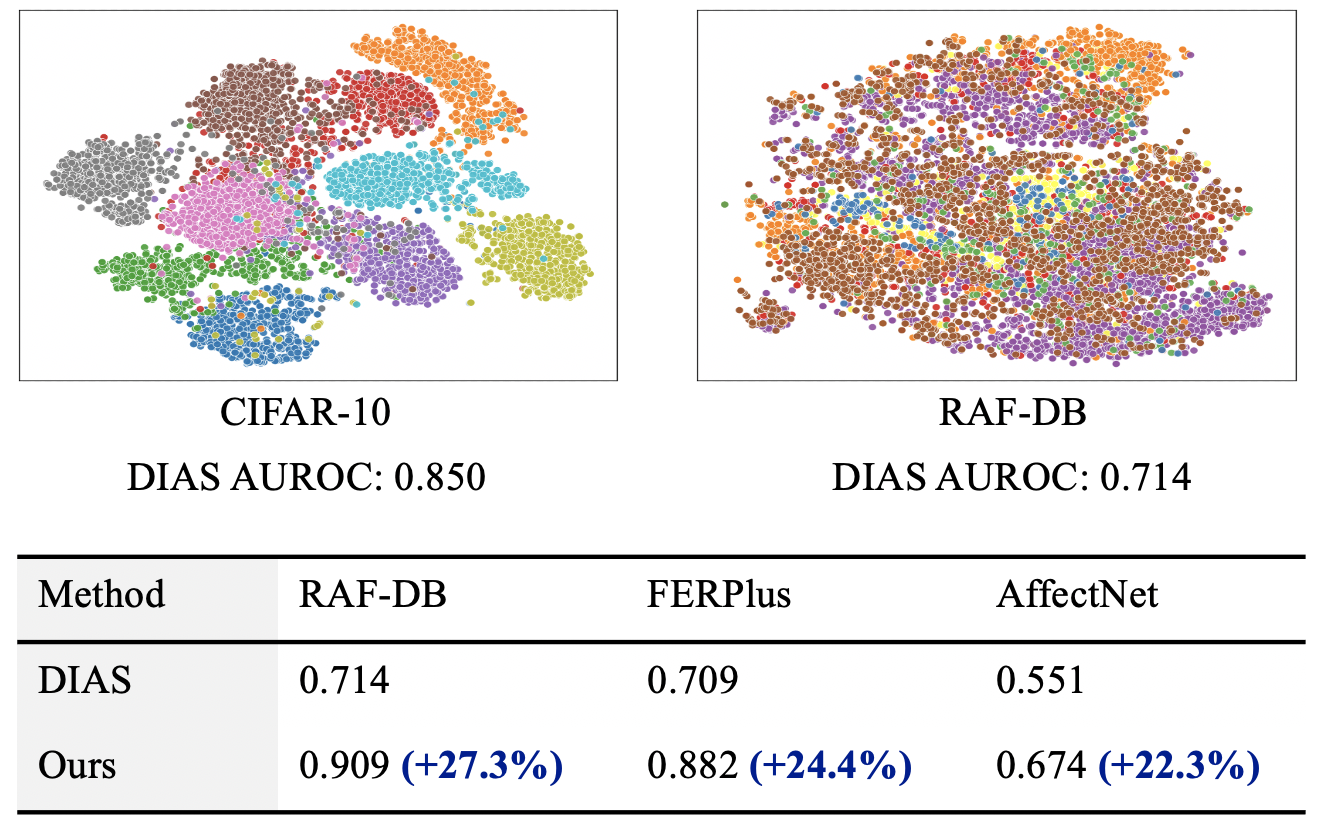}
   \caption{We show the extracted features using CLIP on CIFAR-10 and RAF-DB. CIFAR-10 (RAF-DB) has a large (small) inter-class distance. The small inter-class distance of FER data makes open-set samples similar to close-set samples and degrades the performance of the SOTA open-set recognition method DIAS from 0.850 to 0.714. Our method outperforms DIAS by large margins (over $20\%$ improvement based on the original AUROC) on the open-set FER task of three different FER datasets.}
    \label{illus}
    \vspace{-2mm}
\end{figure}

To solve the above problem, we propose the open-set FER for the first time, which aims to maintain the high accuracy of FER models in the closed set while enabling them to identify samples that belong to unknown classes. Specifically, FER models should be able to detect samples that do not fit perfectly into the close-set classes. However, it is a non-trivial problem because of the overconfidence problem~\cite{nguyen2015deep, goodfellow2014explaining, liu2023enhancing} of deep learning models. While some open-set recognition methods~\cite{zhou2021learning, moon2022difficulty} have tried to solve a similar problem, we argue that they fail in open-set FER. We claim that open-set FER is a much more difficult task because FER data are all human faces with very small inter-class distances. This characteristic makes the open-set samples very similar to the close-set samples, which largely degrades the performance of existing open-set recognition methods.

\begin{figure}[t]
  \centering
  \includegraphics[width=1.\linewidth]{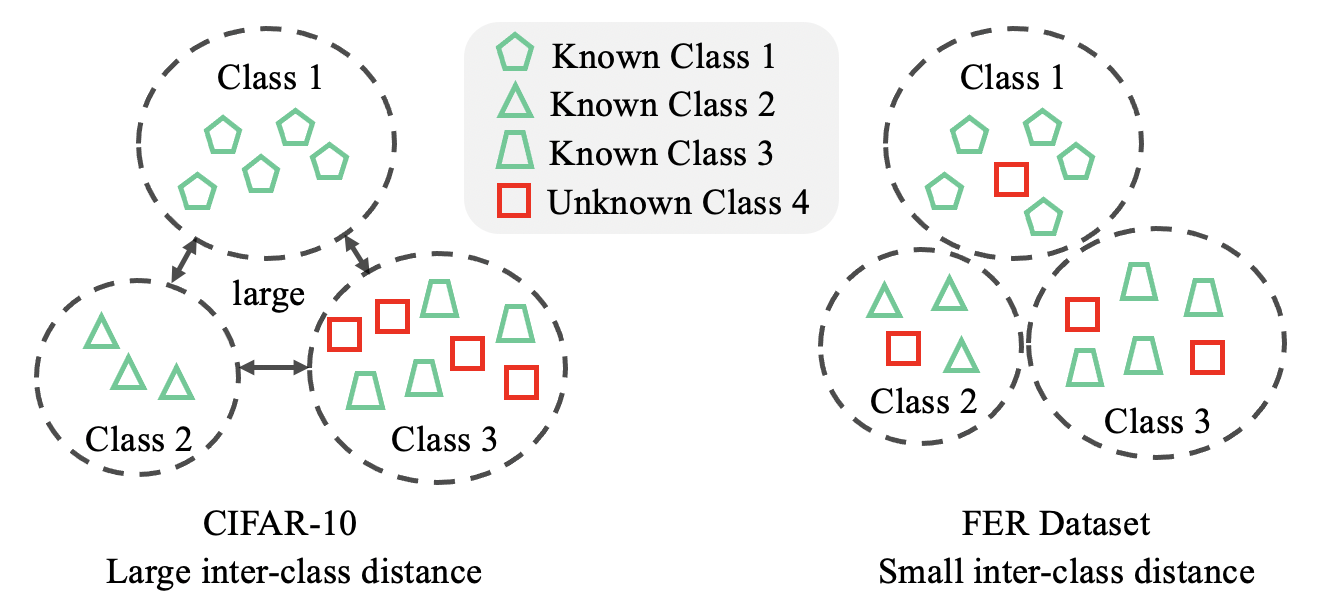}
   \caption{We provide an illustration of our motivation by showing the predicted pseudo labels of the close-set model on CIFAR-10 and FER datasets. CIFAR-10 has relatively large inter-class distances, and the close-set trained model predicts unknown samples into the most similar known class. For example, if the unknown class is 'cat', the trained model will predict almost all cat samples into the known class 'dog'. However, FER data are all human faces. The close-set trained FER model predicts samples of one unknown class to all known classes, which is similar to the concept of symmetric noisy label - a type of easy label noise commonly encountered in the noisy label field.}

    \label{fig1}
\end{figure}

In this paper, different from existing methods, we propose a new way to deal with the open-set FER. Though the small inter-class distance of FER data degrades the existing open-set recognition methods, we turn this characteristic into an advantage. Our motivation is shown in Fig.~\ref{fig1}. We observe that for the data with relatively large inter-class distances like CIFAR-10, the predicted pseudo labels of samples from an unknown class fall into the most semantically similar known class. For example, following the open-set recognition setting~\cite{wang2021energy, zhou2021learning, zhang2023unsupervised}, we consider the 'cat' class as the unknown class and train models on other closed classes. In the test phase, the model predicts almost all unseen 'cat' samples to the known class 'dog'. However, things are different in open-set FER. We observe that the close-set FER model predicts the samples from one unknown class to all known classes. The same phenomenon can easily generalize to the case of several open classes. This strikes us with the concept of symmetric noisy labels~\cite{han2018co} in the noisy label learning field, which is generated by flipping the labels of the original class to all other classes. Symmetric noisy labels are easy to be detected during training as they do not contain much semantic information~\cite{han2018co}. On the contrary, asymmetric label noise which flips the labels of the original class to the most semantically similar class is difficult to be detected. For example, in the CIFAR-10 training phase, if all 'cat' samples are labeled as 'dog' due to noisy labels, then the 'cat' images are very likely to be confidently recognized as 'dog' after training. 
Thus, it is infeasible to transform open-set recognition on CIFAR-10 or other datasets with large inter-class distances into noisy label detection, while it surprisingly works well under the open-set FER task.

Inspired by the aforementioned discussion, we convert open-set FER to a noisy label detection problem for the first time. Unlike existing methods that use softmax scores to detect open-set samples, which tend to be overconfident, we believe that the pseudo labels contain valuable information that has been overlooked. Specifically, we use a close-set trained FER model to predict pseudo labels for all test samples, of which open-set samples will have wrong pseudo labels across all close-set classes. We then introduce cycle training with a cyclically set learning rate inspired by~\cite{huang2019o2u} and iteratively train two FER models to teach each other using the pseudo labels. During training, attention map consistency~\cite{zhang2022learn} is utilized to prevent the model from memorizing the wrong pseudo labels of open-set samples. After training, the loss values of the entire set will form a bimodal distribution, with close-set (open-set) samples having small (large) loss values.

We compare our proposed method with state-of-the-art open-set recognition methods on different open-set FER datasets with different open-set classes. Extensive experiment results show that our method outperforms these methods by large margins. We further show the online prediction ability of our method for one given sample without re-training. More analyses and visualization results of loss values, pseudo labels, and learned features are provided to help further understanding.
 
\begin{itemize}
\item We propose the open-set FER for the first time and observe that existing open-set recognition methods are not effective under this new task due to the small inter-class distance of FER data.
\item Based on our observation of the different distributions of pseudo close-set labels under large and small inter-class distances, we transform open-set FER to noisy label detection for the first time. We further design a novel method with attention map consistency and cycle training to separate close-set and open-set samples.
\item Extensive experiments on different open-set FER datasets show that our method outperforms SOTA open-set recognition methods by large margins. 

\end{itemize}

\begin{figure*}[t]
  \centering
  \includegraphics[width=1.\linewidth]{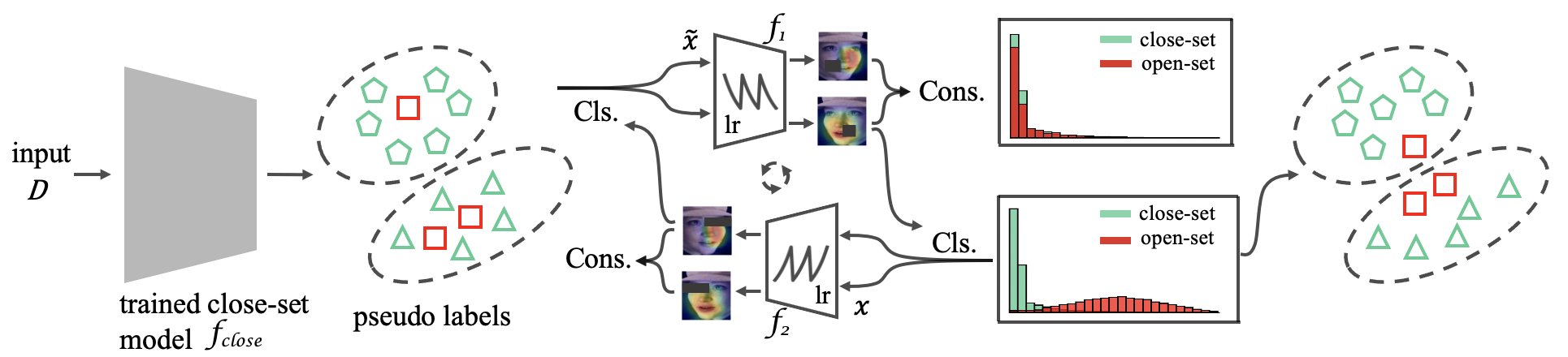}
    \caption{The pipeline of our method. Given the input of both close-set and open-set samples, we utilize the trained close-set model to generate pseudo labels for them. Open-set samples will get noisy close-set labels. We then cyclically train two FER models from scratch with the pseudo labels and utilize attention map consistency loss (Cons.) to prevent the model from memorizing the noisy close-set labels. Each model selects clean samples for another model and teaches each other cyclically. We also utilize a cyclical learning rate (lr) to create an ensemble of models for better separation of close-set and open-set samples. After training, the open-set samples have large classification (Cls.) loss while close-set samples have small Cls. loss.}
    \label{pipe}
    \vspace{-2mm}
\end{figure*}

\section{Related Work}
\label{related work1}
\subsection{Facial Expression Recognition}

Facial expression recognition (FER) is vital for human-computer interaction, and there are numerous studies aimed at improving FER performance~\cite{zhong2012learning, li2017reliable, li2021adaptively, farzaneh2021facial, zhang2021relative, ruan2021feature, li2022towards, zhang2023leave}. For instance, Li~\emph{et al.}~\cite{li2017reliable} use crowd-sourcing to simulate human expression recognition. Farzaneh~\emph{et al.}~\cite{farzaneh2021facial} propose a center loss variant to maximize intra-class similarity and inter-class separation. Ruan~\emph{et al.}~\cite{ruan2021feature} acquire expression-relevant information during the decomposition of an expression feature. Zhang~\emph{et al.}~\cite{zhang2021relative} trains FER models through relative comparison. However, these FER methods are typically evaluated on fixed close-set datasets and they produce highly confident close-set predictions for open-set data.

\subsection{Open-Set Recognition} 
There are two main streams of open-set recognition methods, based on the type of models that they use: discriminative models and generative models \cite{geng2020recent}. The first stream typically employs K+1 way classifiers to discriminate between close-set and open-set data~\cite{scheirer2012toward, bendale2015towards, bendale2016towards, wang2021energy, zhou2021learning}. For example, Bendale \emph{et al.}~\cite{bendale2016towards} replace the softmax layer with OpenMax and calibrate the output probability with Weibull distribution. Wang \emph{et al.}~\cite{wang2021energy} utilize an energy-based K+1 way softmax classifier for open-world setting. Zhou \emph{et al.}~\cite{zhou2021learning} prepare for the unknown classes through learning placeholders for both data and classifier. The second stream of works leverages generative models to generate open-set data and predict the distribution of novel classes~\cite{ge2017generative, neal2018open, oza2019c2ae, perera2020generative, moon2022difficulty}. C2AE~\cite{oza2019c2ae} utilizes class-conditioned auto-encoders and divides the training procedure into two sub-tasks. Perera \emph{et al.}~\cite{perera2020generative} use self-supervision and augment the inputs with generative models to solve the task. The most recent work by Moon \emph{et al.}~\cite{moon2022difficulty} generates open-set samples with diverse difficulty levels to simulate real-world conditions.
While open-set recognition methods have demonstrated good performance when inter-class distances are large, they are not suitable for open-set FER with small inter-class distances. Methods that work well under small inter-class distances are needed.

\section{Problem Definition}
\label{sec:revisit}
Facial expression recognition (FER) models are trained with $\mathcal{D}_{tr}= \left\{ \left( \x _ { i }  , y _ { i }  \right) \right\} _ { i = 1 } ^ {N}$, where
 $\x_{i}$ is a training image and $y_i \in Y  =\{ 1, \ldots , K\}$ is the corresponding label, $N$ and $K$ represent the number of samples and classes. However, in the real-world test set $\mathcal{D}$, there are novel classes, i.e., $\mathcal{D}= \left\{ \left( \x _ { i }  , y _ { i }  \right) \right\} _ { i = 1 } ^ { N }$, $y_i \in \hat{Y}  =\{ 1, \ldots , K , K + 1\}$, where class $K+1$ is the novel category which may contain one or more novel classes. We aim to detect the samples with the label $K+1$, while classifying the samples with label $y_i \in Y$. 
However, as FER data are all human faces, the open-set data is very close to the close-set data. Furthermore, FER models have the overconfidence problem, which makes the confidence scores of both close-set and open-set samples close to~$1$ and drastically degrades the performance of open-set recognition methods. Thus, a more effective method is needed to solve open-set FER.

\section{Method}
We notice that there is useful information from the predicted pseudo labels of the trained close-set model, which have been neglected, shown in Fig.~\ref{fig1}. Unlike the large inter-class distance datasets like CIFAR-10 where most open-set samples are predicted by the trained model into the most similar close-set class, in FER, open-set samples are predicted across all close-set classes, which are similar to the concept of symmetric noisy labels in the noisy label learning field. Symmetric noisy labels are easier to detect than asymmetric noisy labels, which only distribute to the most semantically similar class~\cite{han2018co, kim2019nlnl, zhang2022model} like the pseudo labels in CIFAR-10. Thus, for the first time, we transform the open-set FER into noisy label detection based on the above discussions.

\subsection{Pipeline}
The pipeline of our proposed method is shown in Fig.~\ref{pipe}. Given a pre-trained close-set FER model $f_{close}$, and the input test set $D$, which contains both close-set and open-set samples. We first utilize $f_{close}$ to generate close-set pseudo labels for all the samples in $D$. We then cyclically train two FER models $f_{1}$ and $f_{2}$ from scratch with the close-set pseudo labels. Specifically, given images from $D$, we first apply random erasing~\cite{zhong2020random} to them and the erased images are denoted as $\textbf{x}$. We then flip $\textbf{x}$ to get $\widetilde{\textbf{x}}$. The classification loss is calculated only with the $\textbf{x}$ as \begin{equation}
l_{cls} = -\frac1N \sum_{i=1}^N  (\log{\frac{e^{\textbf{W}_{\textbf{y}_i}f_{1}(\textbf{x}_i)}}{\sum\nolimits_{j}^K e^{\textbf{W}_jf_{1}(\textbf{x}_i)}}}), \label{classification}    
\end{equation}
where $\textbf{W}_{\textbf{y}_i}$ is the $\textbf{y}_i$-th weight from the FC layer, $f_{1}(\textbf{x}_i)$ is the output logits of $\textbf{x}_i$. Inspired by~\cite{zhang2022learn}, we utilize attention map consistency to prevent the model from memorizing the wrong pseudo labels of open-set samples, leading to a large classification loss of open-set samples. The attention maps are computed by multiplying the weight from the FC layer and the features extracted from the last convolution layer following CAM~\cite{zhou2016cvpr}. We denote the attention maps of $\textbf{x}$ and $\widetilde{\textbf{x}}$ as $\textbf{M} \in \mathbb{R}^{N\times K\times H\times W }$ and $\widetilde{\textbf{M}}$, where $N, K, H, W$ represent the number of the samples, expression classes, height and width. The consistency loss is calculated based on the transformation consistency of attention maps according to~\cite{zhang2022learn}, which regularizes the model to focus on the whole facial features and prevents the model from overfitting the wrong pseudo labels of open-set samples.  
\begin{equation} l_{cons} = \frac1{NKHW} \sum_{i=1}^N \sum_{j=1}^K  ||\textbf{M}_{ij} - {Flip(\widetilde{\textbf{M}})_{ij}}||_2.\label{consistency}\end{equation}
The train loss for $f_{1}$ and $f_{2}$ is computed as follows,
\begin{equation}l_{train} = l_{cls} + \lambda l_{cons}.
\label{totalloss}
\end{equation}
We set the consistency weight $\lambda$ as $5$ across all experiments. 
We further introduce cycle training to improve the performance of open-set detection. First, we cyclically set the learning rate as the initial learning rate every $10$ epoch inspired by~\cite{huang2019o2u}, which is similar to an ensemble of several models with different states to help detect noisy (open-set) samples. Second, we cyclically set the training of $f_1$ and $f_2$. Specifically, at the first step, we train $f_1$, then we utilize Gaussian Mixture Models~\cite{li2020dividemix} to model the classification loss of $f_1$ and select the clean samples for the training of $f_2$, we set the threshold of selection as $0.5$. In the second step, $f_2$ selects clean samples for the training of $f_1$. We repeat the two steps until the two models converge. After training, the open-set samples are associated with large classification loss values which can be easily separated from close-set samples with small classification loss values.

\subsection{Novelty and Contribution}
We claim our novelty and contribution as introducing a new FER task and designing a new pipeline, which outperforms SOTA open-set recognition methods by large margins, along with our new discoveries and insights instead of simply introducing a new technical method. We are the first to propose the open-set FER task based on recent works~\cite{cowen2021sixteen, bryant2022multi, kollias2023multi} and find that existing SOTA methods do not perform well under this task. Our discovery that pseudo labels of open-set FER samples are similar to symmetric noisy labels is novel. Inspired by that, we design a new pipeline and propose a new method including cycle training and attention map consistency to address open-set FER from a noisy label detection perspective, which has not been done before. Our approach outperforms SOTA open-set recognition methods by large margins. Though with relatively small technical contribution, we believe that the new discoveries and good performance are our main contributions.

\section{Experiments}
\subsection{Datasets}
RAF-DB \cite{li2017reliable} is annotated with seven basic expressions by 40 trained human coders, including 12,271 images for training and 3,068 images for testing.

FERPlus \cite{barsoum2016training} is extended from FER2013 \cite{goodfellow2013challenges}, which consists of 28,709 training images and 3,589 test images collected by the Google search engine. We utilize the same seven basic classes as RAF-DB in our experiments.

AffectNet \cite{mollahosseini2017affectnet} is a large-scale FER dataset, which contains eight expressions. There are 286,564 training images and 4,000 test images.

\subsection{Implementation Details}
Following open-set recognition setting~\cite{geng2020recent}, open-set samples should be semantically different from close-set samples while they do not have the domain gap. Specifically, we construct close-set and open-set from the above FER datasets. We set some classes as open-set classes and the rest are close-set classes following~\cite{geng2020recent, moon2022difficulty}. For training, we utilize close-set samples of the train set. The test set is the original test set containing both open-set and close-set samples plus the remaining open-set samples of the train set. We utilize ResNet-18 as the backbone. The learning rate $\eta$ is set to $0.0002$ and we use Adam~\cite{kingma2014adam} optimizer with weight decay of $0.0001$. The max training epoch $T_{max}$ is set to 40. As our method does not affect the classification accuracy of the original close-set FER performance, we mainly focus on the detection of open-set data. Close-set classification accuracy of different methods is shown in the Supp. material. We utilize two widely used metrics AUROC~\cite{liang2018enhancing} and FPR@TPR95~\cite{hendrycks2017a} in the open-set recognition field, they both range from 0 to 1. AUROC is the area under the Receiver Operating Characteristic (ROC) curve, the higher the better, while FPR@TPR95 measures the false positive rate (FPR) when the true positive rate (TPR) is 0.95, the lower the better.

\begin{table*}[t]
\begin{center}
\footnotesize
\setlength{\tabcolsep}{1.1mm}
\begin{tabular}{lccccc|ccccc|lccccc}
\toprule[1pt]
\multicolumn{1}{l|}{Metric}           & \multicolumn{5}{c|}{AUROC ($\uparrow$)}                    & \multicolumn{5}{c|}{FPR@TPR95 ($\downarrow$)}         &\multicolumn{1}{l|}{Metric}           & \multicolumn{5}{c}{AUROC ($\uparrow$)}          \\ \midrule
\multicolumn{1}{l|}{Method} & Baseline & EOW   & PROS & DIAS & Ours  & Baseline & EOW & PROS & DIAS & Ours & \multicolumn{1}{l|}{Method} & Baseline & EOW   & PROS & DIAS & Ours \\ \midrule[1pt]
\multicolumn{1}{l|}{R Sur.}         & 0.517    & 0.648 & 0.806 & 0.725     & \textbf{0.918} & 0.926    & 0.897 & 0.730 & 0.850   & \textbf{0.608} & \multicolumn{1}{l|}{F Sur.}   & 0.406    & 0.641 & 0.676 & 0.710      & \textbf{0.933} \\
\multicolumn{1}{l|}{R Fea.}             & 0.411    & 0.577 & 0.706 & 0.660      & \textbf{0.907} & 0.980    & 0.946 & 0.882 & 0.918      & \textbf{0.444} & \multicolumn{1}{l|}{F Fea.}   & 0.370    & 0.581 & 0.664 & 0.634      & \textbf{0.899}\\
\multicolumn{1}{l|}{R Dis.}        & 0.473    & 0.609 & 0.788 & 0.728      & \textbf{0.910} & 0.925    & 0.914 & 0.730 & 0.863      & \textbf{0.462} & \multicolumn{1}{l|}{F Dis.}   & 0.352    & 0.596 & 0.771 & 0.711      & \textbf{0.871} \\
\multicolumn{1}{l|}{R Hap.}          & 0.554    & 0.606 & 0.695 & 0.703      & \textbf{0.892} & 0.852    &0.904& 0.881 & 0.823      & \textbf{0.528} & \multicolumn{1}{l|}{F Hap.}   & 0.476    & 0.645 & 0.731 & 0.726      & \textbf{0.855} \\
\multicolumn{1}{l|}{R Sad.}          & 0.506    & 0.654 & 0.738 & 0.668      & \textbf{0.911} & 0.930    & 0.940 & 0.798 & 0.879      & \textbf{0.519} & \multicolumn{1}{l|}{F Sad.}    & 0.413    & 0.575 & 0.681 & 0.665      & \textbf{0.862}\\
\multicolumn{1}{l|}{R Ang.}            & 0.450    & 0.720 & 0.704 & 0.734      & \textbf{0.906} & 0.937   & 0.814 & 0.877 & 0.857      & \textbf{0.684} & \multicolumn{1}{l|}{F Ang.}   & 0.410    & 0.578 & 0.798 & 0.753      & \textbf{0.891} \\
\multicolumn{1}{l|}{R Neu.}          & 0.566    & 0.606 & 0.803 & 0.778      & \textbf{0.917} & 0.866    & 0.977 & 0.777 & 0.819      & \textbf{0.587} & \multicolumn{1}{l|}{F Neu.}   & 0.544    & 0.529 & 0.852 & 0.767      & \textbf{0.864}\\ \midrule
\multicolumn{1}{l|}{Mean}             & 0.497    & 0.631 & 0.749 & 0.714      & \textbf{0.909} & 0.917   & 0.913 & 0.811 & 0.858      & \textbf{0.547} & \multicolumn{1}{l|}{Mean}  & 0.424    & 0.592 & 0.739 & 0.709      & \textbf{0.882}\\

\bottomrule[1pt]
\end{tabular}
\end{center}
\caption{The detection performance of state-of-the-art open-set recognition methods on open-set FER. We start with the one-class open-set FER and utilize two common metrics AUROC (higher the better) and FPR@TPR95 (lower the better) for evaluation. The expression class listed on the left is the open-set class (Sur.: Surprise, Fea.: Fear, Dis.: Disgust, Hap.: Happiness, Sad.: Sadness, Ang.: Anger, Neu.: Neutral). 'R' represents RAF-DB and 'F' represents FERPlus. 'PROS' is 'PROSER' for short. Our method outperforms the state-of-the-art open-set recognition methods on open-set FER tasks with very large margins.}
\label{experiment1}
\end{table*}

\begin{table}[t]
\begin{center}
\footnotesize
\setlength{\tabcolsep}{1.4mm}
\begin{tabular}{lccccc}
\toprule[1pt]

\multicolumn{1}{l|}{Open class} & Baseline & EOW   & PROSER & DIAS & Ours  \\ \midrule[1pt]
\multicolumn{1}{l|}{Sur.+Fea.}         & 0.436   & 0.561 & 0.763 & 0.706  & \textbf{0.916}\\
\multicolumn{1}{l|}{Fea.+Dis.}             & 0.445 & 0.583 & 0.764 & 0.688     & \textbf{0.884}\\
\multicolumn{1}{l|}{Dis.+Hap.}        & 0.575    & 0.632 & 0.727 & 0.700     & \textbf{0.887} \\
\multicolumn{1}{l|}{Hap.+Sad.}          & 0.609    & 0.536 & 0.726 & 0.717     & \textbf{0.865}\\
\multicolumn{1}{l|}{Sad.+Ang.}          & 0.486    & 0.691 & 0.718 & 0.702      & \textbf{0.879}\\ 
\multicolumn{1}{l|}{Ang.+Neu.}          & 0.552    & 0.634 & 0.825 & 0.769      & \textbf{0.895}\\ \midrule

\multicolumn{1}{l|}{Sur.+Fea.+Dis.}         & 0.497   & 0.588 & 0.769 & 0.731  & \textbf{0.893}\\
\multicolumn{1}{l|}{Fea.+Dis.+Hap.}             & 0.592   & 0.534 & 0.667 & 0.685     & \textbf{0.880}\\
\multicolumn{1}{l|}{Dis.+Hap.+Sad.}        & 0.649    & 0.624 & 0.743 & 0.723     & \textbf{0.815} \\
\multicolumn{1}{l|}{Hap.+Sad.+Ang.}          & 0.637    & 0.698 & 0.705 & 0.718    & \textbf{0.840} \\
\multicolumn{1}{l|}{Sad.+Ang.+Neu.}          & 0.630   & 0.750 & 0.829 & 0.802      & \textbf{0.883}\\ \midrule
\multicolumn{1}{l|}{Mean}             & 0.555    & 0.621 & 0.749& 0.722     & \textbf{0.876}\\
\bottomrule[1pt]
\end{tabular}
\end{center}
\caption{The detection performance (AUROC) of different methods with two or three open classes. Our method achieves the best results under all settings.}
\label{experiment2}
\vspace{-2mm}
\end{table}

\begin{table}[t]
\begin{center}
\footnotesize
\setlength{\tabcolsep}{2.0mm}
\begin{tabular}{l|ccccc}
\toprule[1pt]
Dataset & Baseline & EOW   & PROSER & DIAS  & Ours  \\ \midrule[1pt]
Compound & 0.665    & 0.679 & 0.648  & 0.674 & \textbf{0.771}\\ 
AffectNet  & 0.552    & 0.507 & 0.610  & 0.551 & \textbf{0.674} \\ 
\bottomrule[1pt]
\end{tabular}
\end{center}
\caption{The AUROC of different methods on compound classes and AffectNet. Compound classes are very similar to the basic classes and AffectNet is a large-scale FER dataset with lots of label noises. They both degrade the open-set detection performance, while our method still achieves the best performance.}
\label{compound}
\vspace{-3mm}
\end{table}

\subsection{Open-Set FER With One or Several Basic Classes}
\label{openwithoneclass}

The open-set recognition performance is reported in Table~\ref{experiment1}. The baseline method is MSP~\cite{hendrycks2017a} utilizing softmax score to detect open-set samples. EOW~\cite{wang2021energy}, PROSER(PROS)~\cite{zhou2021learning}, DIAS~\cite{moon2022difficulty} are state-of-the-art open-set recognition methods. Results show that our method not only outperforms all other methods on the mean performance but also achieves the best performance with different open-set classes. Furthermore, the improvements brought by our method are significant. For example, the mean performance of the baseline on the RAF-DB dataset is 0.497 AUROC, which is similar to a random guess. There is AUROC lower than 0.5 as we maintain the range meaning of the softmax score across different experiments, a lower softmax score always means the sample is more like open-set samples. PROSER (discriminative method) and DIAS (generative method) improve the baseline to 0.749 and 0.714, respectively. Our method further improves the AUROC from around 0.7 to around 0.9, which is impressive.

We further carry out experiments to validate the effectiveness of our method when the open-set data contains more than one class with results displayed in Table~\ref{experiment2}. Our method outperforms other methods under all settings. We find that the number of open-set classes has little effect on our method. For instance, our method achieves a mean AUROC of $0.876$ with more than one open class, which is only slightly lower than $0.909$ with one open class.

\subsection{Compound Classes and Different Classes}
In the real-world deployment, FER models will encounter compound expressions, which cannot be simply classified into the basic classes~\cite{du2014compound}. We utilize all basic expression images of RAF-DB as the close-set and all compound expression images of RAF-DB as the open set. Results in Table~\ref{compound} illustrate that the performance of all methods drops compared with open basic classes.  Compound classes usually contain several basic expressions, which are more similar to close-set classes than unseen basic classes. Though detecting compound expressions is harder, our method still achieves the best performance of 0.771 AUROC and outperforms other methods with large margins. 

To simulate the situation when different classes are encountered. We use the seven basic classes of AffectNet as close-set classes and the contempt class of AffectNet as the open-set class. In Table~\ref{compound}, our method reaches the best AUROC of 0.674. The detection performance of all methods drops as the labels of AffectNet are very noisy, leading to a low close-set classification accuracy of around 60\%, which is significantly lower than the accuracy of around 90\% achieved on RAF-DB. As claimed by work~\cite{vaze2022openset}, a good close-set classifier leads to high detection performance, which means the performance drops of all methods on AffectNet are reasonable.

\begin{table}[t]
\begin{center}
\footnotesize
\setlength{\tabcolsep}{3.0mm}
\begin{tabular}{lc|lc}
\toprule[1pt]
Open class   & AUROC       & Open class & AUROC       \\ \midrule[1pt]
Surprise           & 0.918/0.894 & Sadness         & 0.911/0.882 \\
Fear            & 0.907/0.870 & Anger             & 0.906/0.898 \\
Disgust            & 0.910/0.899 & Neutral              & 0.917/0.889 \\
Happiness              & 0.892/0.846 & Mean           & 0.909/0.883 \\ 
\bottomrule[1pt]
\end{tabular}
\end{center}
\caption{Offline/Online detection performance of our method. The offline method achieves better performance, while it is slightly less efficient than the online method. The mean AUROC drops $2.6\%$, which is acceptable, as our online version still achieves better AUROC of $0.883$ compared with other state-of-the-art open-set recognition methods, whose best AUROC is $0.749$.}
\label{online}
\end{table}

\begin{figure}[!h]
  \centering
  \includegraphics[width=1.\linewidth]{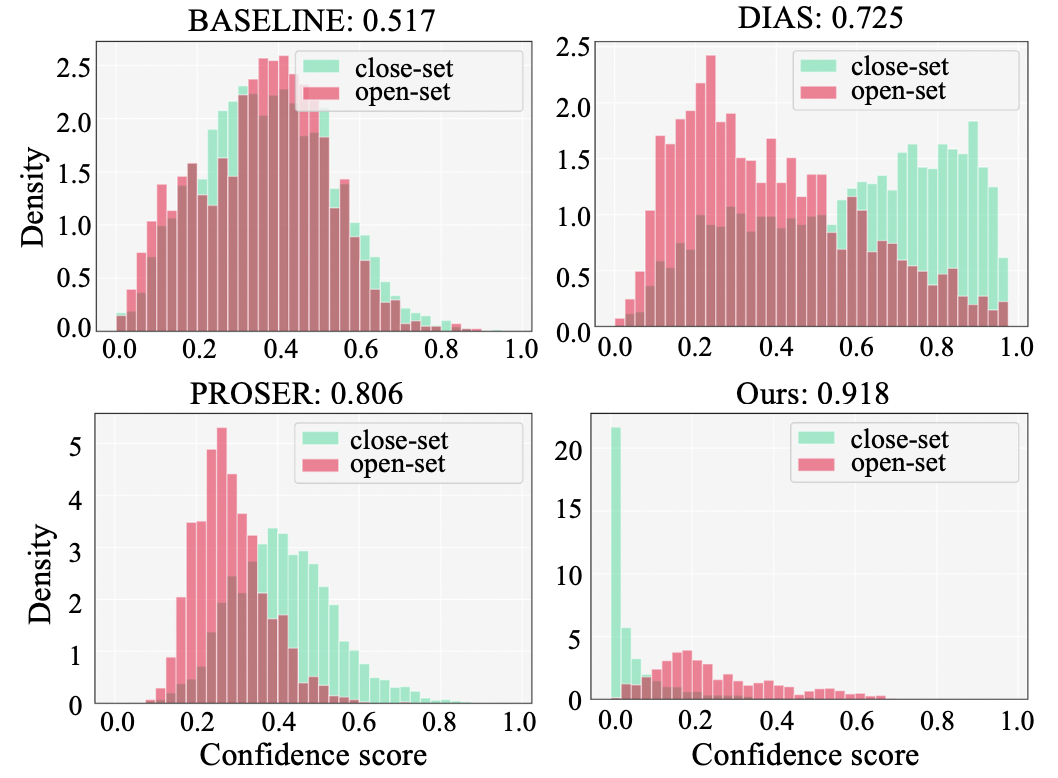}
    \caption{Confidence scores of different methods. AUROC of each method is marked below. The baseline method fails as FER data have small inter-class distances, making open-set data have the same range of confidence scores as close-set data. Close-set and open-set data are separated by DIAS and PROSER while they still overlap a lot. Our method transforms open-set FER to noisy label detection and effectively separates close-set and open-set samples.}
    \label{lossvalue}
    \vspace{-2mm}
\end{figure}

\subsection{Online Application for One Given Sample}
As our method needs to train a model from scratch, one may ask whether our method is suitable for the online detection of only one given test sample at a time. We show that once trained, our method can be utilized for online detection as the test time classification loss can still indicate whether a sample is open-set. The experiment details are in the Supp. material and the results are shown in Table~\ref{online}. We train the model only once and evaluate it on the fly to simulate the online detection of one test sample at a time. The mean AUROC of seven classes drops from 0.909 to 0.883 (2.6\%), which is acceptable. Though with slightly lower AUROC, the online version is more efficient as we do not need to retrain our model each time we encounter new open-set data. Note that the online version achieves an AUROC of $0.883$, which still outperforms other state-of-the-art open-set recognition methods as their best AUROC is $0.749$ in Table~\ref{experiment1}.

\section{Further Analyses}

\subsection{Visualization of Confidence Scores}
We visualize the confidence scores, which are utilized to detect open-set data in Fig.~\ref{lossvalue}. The confidence score of our method is the classification loss value. We normalize confidence scores to $[0, 1]$ to make comparisons. The results in Fig.~\ref{lossvalue} demonstrate that when no open-set recognition methods are used, the confidence scores of close-set and open-set samples overlap considerably. Although open-set recognition methods such as DIAS (generative) and PROSER (discriminative) perform better than baseline, they still have significant overlap. In contrast, our method achieves the best performance and effectively separates close-set and open-set samples. This is because we view open-set FER from a unique perspective of noisy label detection. By utilizing the useful information from pseudo labels, which implicitly encode the information of close/open set, we are able to mitigate the overconfidence problem.

\begin{table}[]
\begin{center}
\footnotesize
\setlength{\tabcolsep}{2.5mm}
\begin{tabular}{ccc|c}
\toprule[1pt]
Attention & Cycle LR & Cycle Training& AUROC \\ \midrule[1pt]
     &            &        & 0.517  \\
 \checkmark    &  &        &   0.885   \\
 \checkmark & \checkmark & &  0.912   \\
 \checkmark & \checkmark &\checkmark & 0.918\\ 
 \bottomrule[1pt]
\end{tabular}
\end{center}
\caption{The ablation study of our method.}
\label{ablation}
\end{table}

\begin{figure}[t]
  \centering
  \includegraphics[width=1.\linewidth]{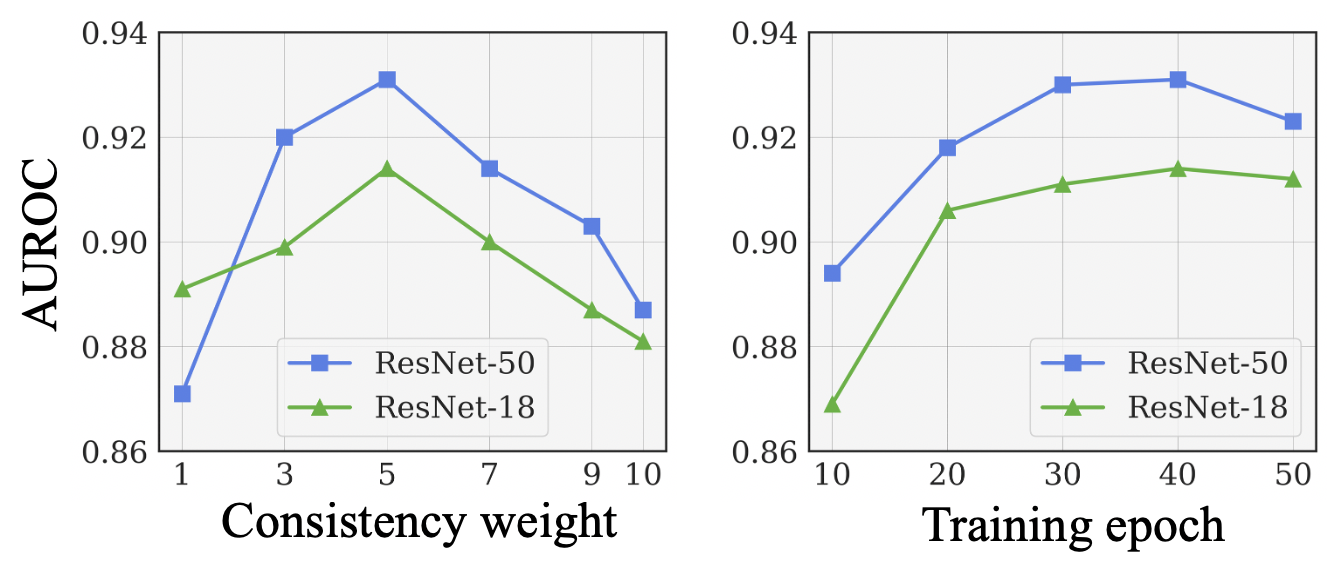}
    \caption{Hyperparameters analyses of our method on ResNet-$18$ and ResNet-$50$. ResNet-$50$ generally has better performance than ResNet-$18$. Our method is not sensitive to hyperparameters as AUROC slightly changes from $0.87$ to $0.93$. The best consistency weight is $5$ and the best training epoch number is $40$.}
    \label{hyper}
    \vspace{-2mm}
\end{figure}

\subsection{Ablation and Hyperparameter Study}
To show the effectiveness of each of the modules in our method, we carry out an ablation study on RAF-DB utilizing surprise as the open-set class. The AUROC in Table~\ref{ablation} illustrates that the most effective module in our method is the attention map consistency (Attention), which can prevent the FER model from memorizing the wrong pseudo labels of open-set samples. Cycle learning rate (LR) improves the performance by setting the learning rate cyclically to make an ensemble to detect wrong pseudo labels~\cite{huang2019o2u}. Cycle training further utilizes two FER models to select clean samples to iteratively teach each other for better performance. Each of the introduced modules contributes to the good performance of our method.

Our method has two main hyperparameters, which are the weight of consistency loss and the training epoch number. We carry out experiments with the consistency weight ranges from 1 to 10 and training epoch number ranges from 10 to 50, as shown in Fig.~\ref{hyper}. Overall, our method is not sensitive to the two hyperparameters as the AUROC only changes from 0.87 to 0.93 under two different backbones and all different hyperparameters. ResNet-50 performs better than ResNet-18. However, in order to fairly compare with other methods, we report the performance using ResNet-18 in Table~\ref{experiment1}. As for the consistency weight, the performance increases and then decreases as a small consistency weight is not enough to prevent the model from memorizing the open-set (noisy) samples and a very large consistency weight impedes the optimization of classification loss. We set the consistency weight as 5 in all our experiments. Training epoch number has little effect on the performance. We simply set the training epoch number as 40 in all our experiments.

\begin{figure}[t]
  \centering
  \includegraphics[width=1.\linewidth]{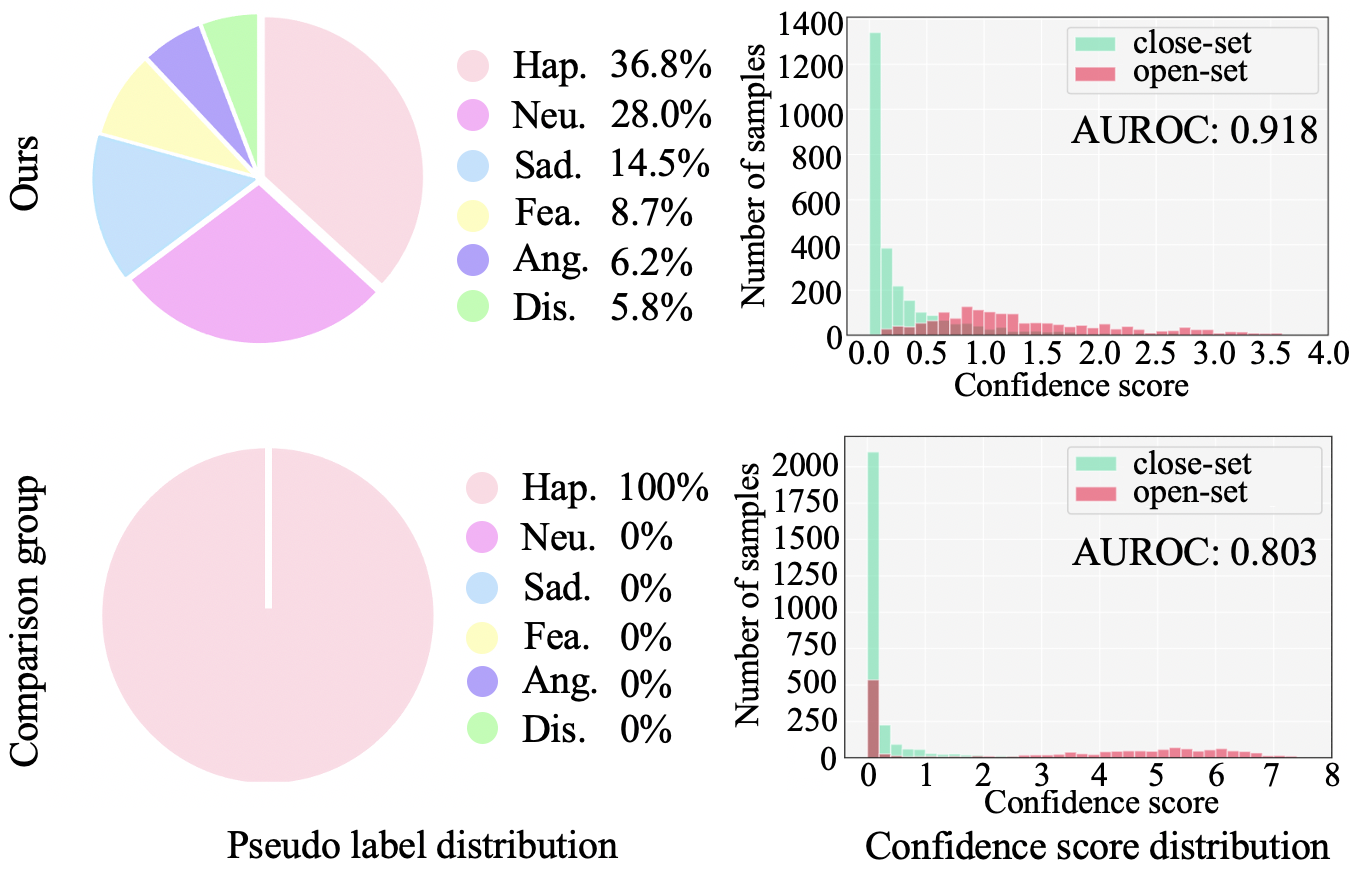}
    \caption{We design a comparison group to study the effect of the pseudo-label distribution. Our motivation is that the pseudo labels of open-set samples distribute across all close-set classes.  The comparison group has the pseudo labels of open samples belonging to the most similar close-set class. The detection performance drops from 0.918 to 0.803. This illustrates that our method works well because the pseudo labels distribute across all close-set classes instead of centering to the class with the largest number of pseudo labels (happiness in the comparison group).}
    \label{pseudo} 
    \vspace{-2mm}
\end{figure}

\begin{figure}[t]
  \centering
  \includegraphics[width=1.\linewidth]{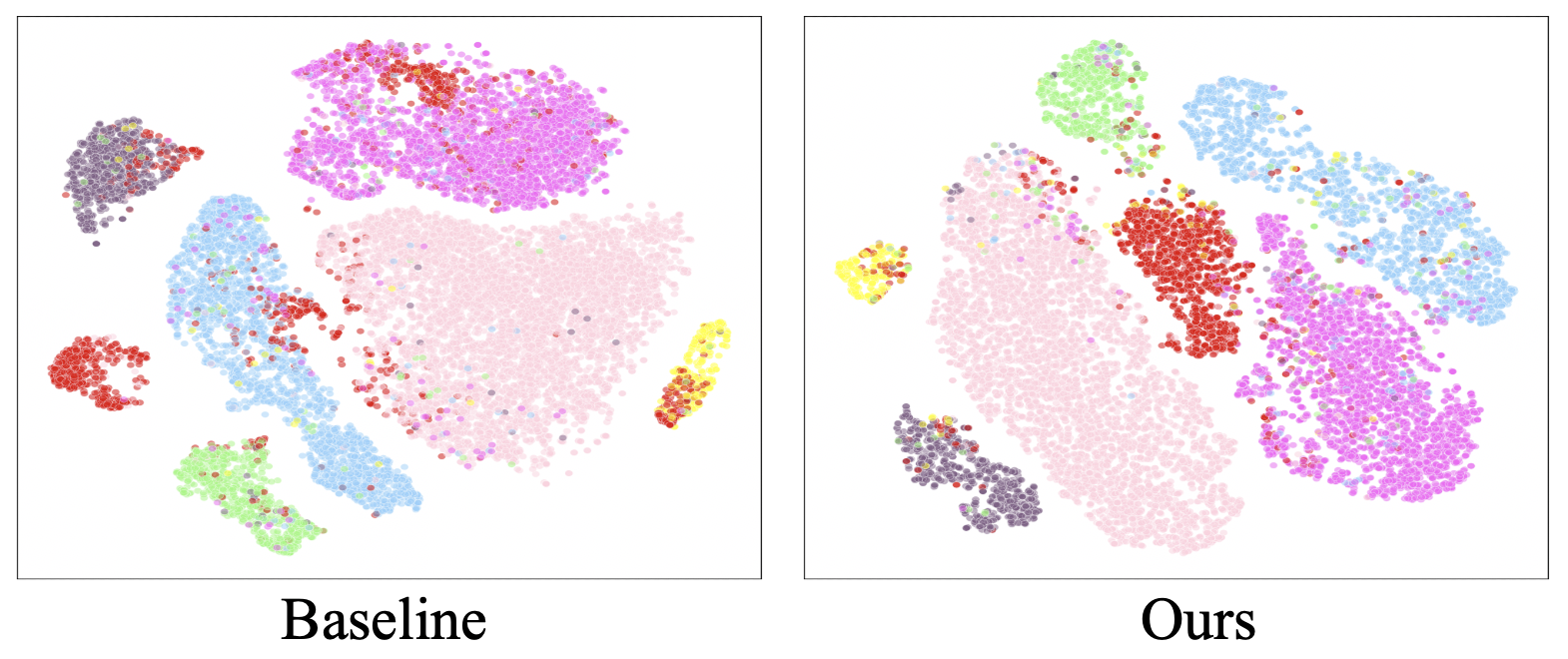}
    \caption{The learned features of baseline and our method. The features are shown with the latent truth. \emph{Open-set features are marked as red.} The learned open-set features of the baseline method are mixed with close-set features, while our method does not overfit the wrong pseudo labels of open-set samples and separates open-set features from close-set features.}
    \label{tsne}
    \vspace{-2mm}
\end{figure}

\subsection{Analyses of Pseudo Label Distributions}
To dig deeper into why our method works well in open-set FER, we provide more analysis of the pseudo labels. We argue that our method is effective because it utilizes information from the pseudo labels which are neglected by previous methods. We first plot the distribution of the predicted pseudo labels of open-set samples in Fig.~\ref{pseudo}. We observe that the pseudo labels of the surprise (open-set) class are distributed across all close-set classes. To exclude the influence of the happiness class with the largest number of pseudo labels, we design a comparison group with all the pseudo labels of open-set (surprise) samples lying in the happiness class. Shown in Fig.~\ref{pseudo}, the open-set recognition performance drops from $0.918$ to $0.803$. We observe that though some open-set samples are correctly detected, there are many open-set samples confused with close-set samples. The results illustrate that semantically similar label noise, e.g., labeling all surprise samples to happiness, is harder to detect. They also demonstrate that our method works well because the pseudo labels distribute across all classes instead of centering on one class with the largest number of pseudo labels.

\subsection{Visualization of Learned Features}
We utilize t-SNE~\cite{van2008visualizing} to visualize the learned features. The results are shown in Fig.~\ref{tsne}, which implies that the baseline method memorizes the wrong pseudo labels and the learned open-set features (marked as red) are mixed with other close-set features after training.
However, our method prevents the model from memorizing the wrong labels of open-set samples, which avoids pushing open-set features into the close-set feature clusters. After training, the model learns useful features from the given samples and set the open-set features apart from the other close-set features.

\section{Conclusion}
\label{conclusion}
We propose a new topic named open-set facial expression recognition (FER) to address the overconfidence problem of FER models and maintain their ability to reject unseen samples. As FER data have small inter-class distances, existing open-set recognition methods are not well suited for open-set FER. However, we find that due to this characteristic, the pseudo labels assigned to open-set samples are distributed across all close-set classes, which is similar to the concept of symmetric noisy labels. Inspired by that, we propose a novel method to convert open-set FER into a noisy label detection problem. Utilizing extra information of pseudo labels and together with cycle training and attention map consistency, our method gets rid of the overconfidence problem of softmax scores and effectively detects the open-set samples. Extensive experiments on different open-set FER datasets and open-set classes show that our method outperforms state-of-the-art open-set recognition methods by large margins. We believe that our work will enlighten more research works on the relationship between the open-set recognition field and the noisy label detection field.

\section{Acknowledgments}
We sincerely thank SPC who find our work valuable and the reviewers who have given us lots of valuable suggestions. This work was supported in part by the National Natural Science Foundation of China under Grant No.62276030 and 62236003, in part by the BUPT Excellent Ph.D. Students Foundation No.CX2023111 and in part by scholarships from China Scholarship Council (CSC) under Grant CSC No.202206470048.

\bibliography{anonymous-submission-latex-2023}

\end{document}